\documentclass[conference]{IEEEtran}
\IEEEoverridecommandlockouts

\usepackage{cite}
\usepackage{enumitem}
\usepackage{amsmath,amssymb,amsfonts}
\usepackage{algorithmic}
\usepackage{graphicx}
\usepackage{textcomp}
\usepackage{xcolor}

\usepackage{microtype}      
\usepackage{lipsum}
\usepackage{graphicx}
\usepackage{tcolorbox}
\usepackage{tabularx}

\usepackage{forest}
\usetikzlibrary{trees,positioning,shapes,shadows,arrows.meta}
\usepackage{tikz}

\makeatletter
\newcommand{\linebreakand}{%
  \end{@IEEEauthorhalign}
  \hfill\mbox{}\par
  \mbox{}\hfill\begin{@IEEEauthorhalign}
}
\makeatother

\begin{document}

\title{The Cat and Mouse Game: The Ongoing Arms Race Between Diffusion Models and Detection Methods}


\author{
    \IEEEauthorblockN{
        Linda Laurier\textsuperscript{1}, 
        Ave Giulietta \textsuperscript{2},
        Arlo Octavia \textsuperscript{3}
        Meade Cleti\textsuperscript{*,4}
    }
    \IEEEauthorblockA{
        \textsuperscript{1}Hampton College, USA
    }
    \IEEEauthorblockA{
        \textsuperscript{2}Texas A\&M University, USA
    }
    \IEEEauthorblockA{
        \textsuperscript{3}Liberty University, USA
    }
    \IEEEauthorblockA{
        \textsuperscript{4}Arizona State University, USA
    }
    \IEEEauthorblockA{
        *Corresponding Email: mcleti@asu.edu
    }
}

\maketitle

\begin{IEEEkeywords}
Diffusion Models, Generative Artificial Intelligence, Deepfake Detection, Synthetic Media Detection, AI-generated Content
\end{IEEEkeywords}

\begin{abstract}
The emergence of diffusion models has transformed synthetic media generation, offering unmatched realism and control over content creation. These advancements have driven innovation across fields such as art, design, and scientific visualization. However, they also introduce significant ethical and societal challenges, particularly through the creation of hyper-realistic images that can facilitate deepfakes, misinformation, and unauthorized reproduction of copyrighted material. In response, the need for effective detection mechanisms has become increasingly urgent. This review examines the evolving adversarial relationship between diffusion model development and the advancement of detection methods. We present a thorough analysis of contemporary detection strategies, including frequency and spatial domain techniques, deep learning-based approaches, and hybrid models that combine multiple methodologies. We also highlight the importance of diverse datasets and standardized evaluation metrics in improving detection accuracy and generalizability. Our discussion explores the practical applications of these detection systems in copyright protection, misinformation prevention, and forensic analysis, while also addressing the ethical implications of synthetic media. Finally, we identify key research gaps and propose future directions to enhance the robustness and adaptability of detection methods in line with the rapid advancements of diffusion models. This review emphasizes the necessity of a comprehensive approach to mitigating the risks associated with AI-generated content in an increasingly digital world.
\end{abstract}

\section{Introduction}

The rapid advancement of diffusion models represents a pivotal shift in synthetic media generation. These models offer an unparalleled degree of control and realism, outpacing GANs in producing high-quality, diverse images \cite{Luo2024LaRE, Wang2020CNN}. Platforms like Midjourney and Stable Diffusion have made this technology widely accessible, enabling users, even without technical expertise, to generate photorealistic content from simple text prompts \cite{Bammey2023Synthbuster}. This democratization of content creation fosters innovation in various fields. For example, in art and design, diffusion models are used to explore new aesthetic possibilities \cite{Song2024Trinity}, while in fields such as medical imaging and scientific visualization, they assist in generating highly detailed and accurate visual data for analysis \cite{Pinaya2022Fast, niu2024text}.

However, the increasing sophistication and accessibility of diffusion models also give rise to significant ethical and societal concerns. Their capacity to generate hyper-realistic images, including the ability to synthesize visuals from textual descriptions \cite{Sha2022DE}, opens the door to malicious uses. Deepfakes, for instance, can be weaponized to manipulate public opinion and spread misinformation at an unprecedented scale \cite{Das2023Universal, Papa2023On}. Additionally, the widespread use of these models raises serious copyright and intellectual property issues, as diffusion models can inadvertently reproduce content from their training datasets, raising concerns about unauthorized replication of protected works \cite{Somepalli2023Diffusion, Somepalli2023Understanding, cleti_jano_2024, peng2024securing}. These challenges necessitate the development of robust detection mechanisms to safeguard against the misuse of this powerful technology.

The exceptional realism of images generated by diffusion models threatens the credibility of digital visual media. As these synthetic images become nearly indistinguishable from genuine photographs \cite{T2019Fourier}, the risk of malicious use, including the spread of fake news, creation of fraudulent content, and impersonation, grows exponentially \cite{Deng2023Diffusion, Ricker2022Towards}. Current detection techniques, primarily designed for GAN-generated content, often fail to accurately identify the subtle artifacts and nuanced manipulations characteristic of diffusion-based generation \cite{Wang2023DIRE, Lorenz2023Detecting}.

Furthermore, the rapid evolution of diffusion models, with frequent changes in architectures, training data, and post-processing techniques, demands detection systems that can adapt to new, unseen models. These systems must not only be accurate but also robust to variations in model design and capable of generalizing across different diffusion models \cite{Wang2020CNN, Santosh2024Robust}. The growing prevalence of mixed-content imagery, such as inpainted or subtly altered photos, adds another layer of difficulty to the detection process, as synthetic elements become even harder to distinguish from real ones \cite{Epstein2023Online}. Additionally, diffusion-based text-to-image generation introduces further challenges, complicating the detection of AI-generated text embedded within images \cite{Sha2022DE}.

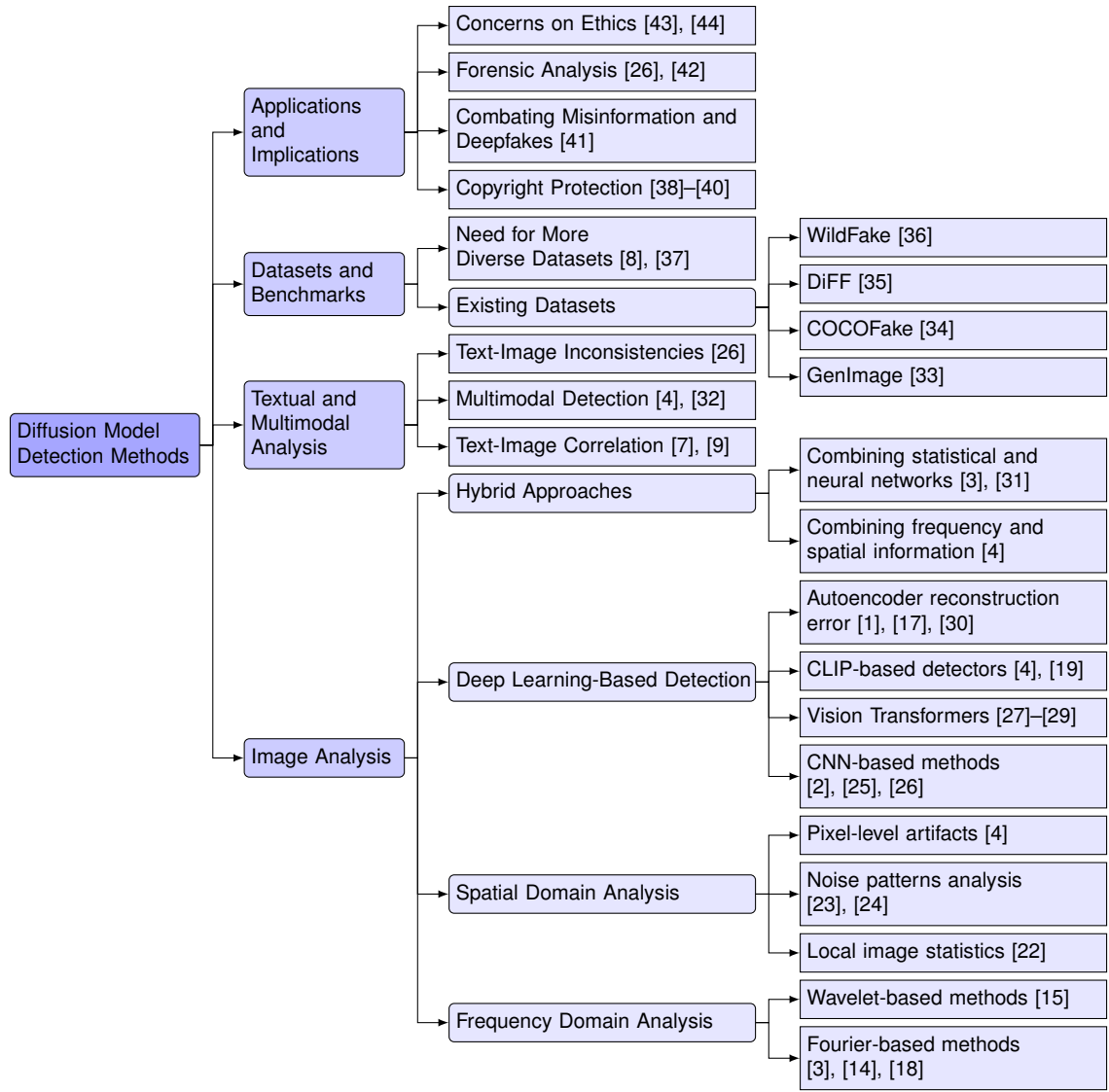
\begin{figure*}
    \centering
    
\tikzset{
    basic/.style  = {draw, align=left, font=\sffamily, rectangle},
    root/.style   = {basic, rounded corners=2pt, thin, fill=blue!35, text width=3cm},
    level1/.style = {basic, thin, rounded corners=2pt, fill=blue!20, text width=2.5cm},
    level2/.style = {basic, thin, rounded corners=2pt, fill=blue!10, text width=5cm},
    leaf/.style   = {basic, thin, fill=blue!10, text width=5cm},
    edge from parent/.style={draw=black, edge from parent fork right},
    level distance=1.0cm,
}

\begin{forest}
for tree={
    grow=east,
    scale=0.8,
    growth parent anchor=west,
    parent anchor=east,
    child anchor=west,
    l sep=6mm,
    s sep=1mm,
    edge path={
        \noexpand\path[\forestoption{edge},->, >={latex}] 
             (!u.parent anchor) -- +(5pt,0pt) |-  (.child anchor) 
             \forestoption{edge label};
    },
    align=left,
}
[{Diffusion Model \\ Detection Methods}, root,
    [{Image Analysis}, level1,
        [{Frequency Domain Analysis}, level2,
            [{Fourier-based methods\\\cite{Bammey2023Synthbuster, T2019Fourier, Lorenz2023Detecting}}, leaf]
            [{Wavelet-based methods \cite{Deng2023Diffusion}}, leaf]
        ]
        [{Spatial Domain Analysis}, level2,
            [{Local image statistics \cite{Wong2023Local}}, leaf]
            [{Noise patterns analysis\\\cite{Liu2022Detecting, Chen2024Single}}, leaf]
            [{Pixel-level artifacts \cite{Song2024Trinity}}, leaf]
        ]
        [{Deep Learning-Based Detection}, level2,
            [{CNN-based methods\\\cite{Wang2020CNN, Coccomini2023Detecting, Sinitsa2024Deep}}, leaf]
            [{Vision Transformers \cite{Aghasanli2023Interpretable, Cozzolino2023Raising, Liu2024Mixture}}, leaf]
            [{CLIP-based detectors \cite{Santosh2024Robust, Song2024Trinity}}, leaf]
            [{Autoencoder reconstruction\\error \cite{Ricker2024AEROBLADE, Wang2023DIRE, Luo2024LaRE}}, leaf]
        ]
        [{Hybrid Approaches}, level2,
            [{Combining frequency and\\spatial information \cite{Song2024Trinity}}, leaf]
            [{Combining statistical and\\neural networks \cite{Ma2023Exposing, Bammey2023Synthbuster}}, leaf]
        ]
    ]
    [{Textual and\\Multimodal\\Analysis}, level1,
        [{Text-Image Correlation \cite{Sha2022DE, Papa2023On}}, leaf]
        [{Multimodal Detection \cite{Song2024Trinity, Xu2023Exposing}}, leaf]
        [{Text-Image Inconsistencies \cite{Sinitsa2024Deep}}, leaf]
    ]
    [{Datasets and\\Benchmarks}, level1,
        [{Existing Datasets}, level2,
            [{GenImage \cite{Zhu2023GenImage}}, leaf]
            [{COCOFake \cite{Amoroso2023Parents}}, leaf]
            [{DiFF \cite{Cheng2024Diffusion}}, leaf]
            [{WildFake \cite{Hong2024WildFake}}, leaf]
        ]
        [{Need for More\\Diverse Datasets \cite{Das2023Universal, Saberi2023Robustness}}, leaf]
    ]
    [{Applications\\and\\Implications}, level1,
        [{Copyright Protection \cite{Wen2023Tree, Min2024Watermark, Hu2024Stable}}, leaf]
        [{Combating Misinformation and\\Deepfakes \cite{Frank2023Representative}}, leaf]
        [{Forensic Analysis \cite{Sinitsa2024Deep, Yu2021Artificial}}, leaf]
        [{Concerns on Ethics \cite{Gandikota2023Erasing, Knott2023Generative}}, leaf]
    ]
]
\end{forest}
    \caption{Taxonomy of Detection Techniques for Diffusion Models, Including Image and Textual Analysis Methods, Datasets, and Applications}
    \label{fig:taxonomy}
\end{figure*}

This articles provides a comprehensive analysis of current research aimed at detecting content generated by diffusion models (see \textbf{Fig \ref{fig:taxonomy}} for the taxonomy). It examines the unique characteristics of diffusion-generated content, such as the subtle artifacts and intricate visual manipulations, and the specific challenges these pose for detection. Additionally, it reviews a wide range of detection methodologies proposed in recent literature, categorizing them by their core techniques, including image analysis, textual analysis (particularly for text-to-image generation), and watermarking or fingerprinting methods. The framework also evaluates existing datasets and benchmarking metrics, stressing the urgent need for more diverse and representative datasets that accurately reflect real-world diffusion model applications \cite{Lin2024Detecting}. Such datasets are essential for ensuring the reliability and effectiveness of detection methods across different domains and use cases.

\section{Fundamentals of Diffusion Models and Detection Challenges}

\subsection{Diffusion Model Content Generation}

Diffusion models generate content by progressively reversing a noise-adding process. Initially, a real image is corrupted step-by-step by adding Gaussian noise over multiple iterations until it becomes indistinguishable from pure noise. The model learns to reverse this process, denoising the image at each step, eventually reconstructing a clean, high-quality synthetic image from random noise \cite{Somepalli2023Diffusion}. Latent Diffusion Models (LDMs) improve the efficiency of this process by performing denoising in a compressed latent space, leveraging a pre-trained autoencoder \cite{Ricker2024AEROBLADE}. Text-to-image diffusion models further complicate the process by incorporating text prompts, aligning generated images with input text, which adds a challenge to detection methods \cite{Coccomini2023Detecting, Sha2022DE}.

\subsection{Unique Characteristics of Diffusion-Generated Content}

Despite their photorealistic appearance, diffusion-generated images exhibit unique characteristics that can assist in their detection. One such feature is \textit{frequency domain artifacts}. Analyzing diffusion-generated images in the Fourier domain often reveals distinct patterns, particularly in high-frequency components \cite{Bammey2023Synthbuster}. Diffusion models tend to underrepresent high frequencies, resulting in noticeable spectral irregularities due to the optimization objectives during training \cite{Ricker2022Towards}. Wavelet-based analysis can also be employed to detect subtle frequency-domain clues \cite{Deng2023Diffusion}.

Another important cue is the presence of \textit{spatial inconsistencies}. Diffusion models often produce images with unusual noise patterns or localized statistical anomalies, which can help distinguish them from real, camera-captured images \cite{Wong2023Local}. These inconsistencies are particularly evident when analyzing pixel relationships in regions with complex textures \cite{Nan2023PatchCraft}. Additionally, \textit{autocorrelation analysis} can reveal anomalous patterns. By measuring correlations between the original image and its shifted versions, researchers can identify deviations that are characteristic of diffusion-generated images \cite{Corvi2023Intriguing}.

Further aiding in detection is the identification of \textit{model-specific fingerprints}. Each diffusion model leaves behind a unique signature in the images it generates, influenced by factors such as architecture, training data, and specific implementation choices. These fingerprints can be applied for both detection and attribution \cite{Das2023Universal}. Techniques like Deep Image Fingerprint have been developed to capitalize on these traits, helping trace the lineage of generated images \cite{Sinitsa2024Deep}.

One of the most frequently observed features in diffusion-generated images is the \textit{underestimation of high frequencies}, leading to less detail and sharpness compared to real images. This underrepresentation is a key target for detection methods, especially in domains like talking face generation, where the lack of high-frequency detail can be particularly noticeable \cite{Ricker2022Towards, Stypukowski2024Diffused}.

\subsection{Challenges in Detecting Diffusion-Generated Content}

One of the central challenges in detecting diffusion-generated content is the \textit{generalization across different diffusion models}. Detectors trained on a single diffusion model often fail when applied to images generated by other models, due to the presence of unique model-specific fingerprints \cite{Wang2020CNN}. This issue is exacerbated by the continuous release of new models, each introducing different variations in output \cite{Epstein2023Online}.

Another major challenge is achieving \textit{robustness to image transformations}. Real-world images undergo numerous transformations such as compression and resizing, which can degrade detection accuracy. Many current detection methods are sensitive to these transformations, limiting their effectiveness in practical applications \cite{Saberi2023Robustness}. Improving robustness to such alterations is an active area of research \cite{Das2023Universal}.

As diffusion models continue to advance, the \textit{subtle differences between real and synthetic images} are becoming more difficult to detect. Sophisticated post-processing techniques, aimed at enhancing the realism of synthetic content, further blur the distinction between real and generated images \cite{Chakraborty2023On, Chen2023On}. This requires the development of more sophisticated detection techniques.

\textit{Detection in mixed-media content} presents additional challenges, especially when synthetic and real content are combined within the same image. For instance, inpainted areas or manipulated sections may go unnoticed without specialized detection methods. Researchers are investigating weakly-supervised localization techniques to address these issues \cite{Epstein2023Online, Tantaru2024Weakly}.

The detection of diffusion-generated content is further complicated in \textit{real-world scenarios}, such as images shared on social media. These images are often subjected to multiple layers of processing, such as compression, which further hinders detection \cite{Papa2023On}. Datasets like WildFake are being developed to simulate real-world conditions, enabling better evaluation of detection methods under practical constraints \cite{Hong2024WildFake}.

Another emerging challenge is \textit{detecting content replication from training data}. Diffusion models may inadvertently replicate content from their training datasets, raising concerns regarding copyright infringement and data misuse \cite{Somepalli2023Diffusion}. Detecting these replications and mitigating such risks is becoming a critical focus of research, with strategies like caption randomization and data augmentation being explored \cite{Somepalli2023Understanding}.

Lastly, specialized fields, such as the detection of \textit{deepfakes of human faces} and \textit{handwriting}, are also under active investigation. Diffusion-generated faces are highly realistic, and detecting these deepfakes remains a particularly difficult task. Specialized datasets, such as DiFF, support research in this area by providing high-quality, realistic deepfake samples for training and evaluation \cite{Deng2023Diffusion, Cheng2024Diffusion}. Similarly, diffusion-generated handwriting presents a new challenge for forgery detection, requiring novel techniques to address this issue \cite{Carriere2023Beyond}.

\section{Detection Methods Based on Image Analysis}

\subsection{Frequency Domain Analysis}

Several studies have used frequency domain characteristics to distinguish diffusion-generated images from real ones. A prominent observation is the challenge diffusion models face in replicating high-frequency details accurately. \cite{T2019Fourier} highlighted the systematic shortcomings of deep network-generated images in replicating high-frequency Fourier modes, which has become a foundational observation for many detection methods.

Building on this, \cite{Bammey2023Synthbuster} introduced a method that analyzes frequency artifacts in the Fourier transform of residual images, demonstrating effectiveness even under mild JPEG compression. However, \cite{Chandrasegaran2021Closer} noted that relying solely on high-frequency discrepancies may be fragile, as minor architectural changes to generative models can mitigate these telltale signs. Moreover, \cite{Lorenz2023Detecting} explored local intrinsic dimensionality, a concept tied to frequency characteristics, employing multi Local Intrinsic Dimensionality (multiLID) for both detection and generator identification.

Research has also examined broader spectral power distribution discrepancies beyond high-frequency components. \cite{Corvi2023Intriguing} systematically analyzed various generators, finding significant differences in mid-to-high-frequency signal content between real and synthetic images. These differences were observable through radial and angular spectral power distributions, suggesting that a more comprehensive spectral analysis can enhance detection.

Wavelet transforms, which analyze images in both frequency and spatial domains, also offer a powerful approach. \cite{Deng2023Diffusion} proposed a multi-scale network that uses wavelet lifting and wavelet-spatial transformer blocks for detecting face forgeries. This method decomposes images into different frequency bands and fuses the resulting features, proving highly robust to various manipulations.

\subsection{Spatial Domain Analysis}

In the spatial domain, researchers have focused on analyzing local image statistics and noise patterns. \cite{Wong2023Local} demonstrated that local statistical properties, which vary across regions of an image, are more effective than global statistics in distinguishing between real and diffusion-generated images. This method also showed robustness to common perturbations.

Noise patterns, both in spatial and frequency domains, offer another line of investigation. \cite{Liu2022Detecting} proposed a method analyzing noise patterns in the frequency domain, finding distinct differences between real and generated images. Similarly, \cite{Chen2024Single} examined noise patterns in small image patches, arguing that generative models often overlook subtle noise characteristics while prioritizing realistic textures in more complex regions.

Pixel-level artifacts further contribute to detection accuracy. \cite{Song2024Trinity} introduced the MCAF unit, which is sensitive to pixel-level artifacts and spectral inconsistencies. This method combines text and pixel features for a more comprehensive detection strategy.

\subsection{Deep Learning-Based Detection}

Deep learning-based techniques have been widely employed for detecting diffusion-generated images. Convolutional Neural Networks (CNNs) remain popular, with \cite{Wang2020CNN} demonstrating that a CNN trained on a single GAN generator could generalize to other CNN-based generators. Traditional CNNs, such as those described in \cite{Coccomini2023Detecting}, continue to be effective for detection tasks, while \cite{Sinitsa2024Deep} used CNN architectural properties for detection and model lineage analysis.

Vision Transformers (ViTs) provide an alternative to CNNs. For instance, \cite{Aghasanli2023Interpretable} combined fine-tuned ViTs with SVMs for deepfake detection, while \cite{Cozzolino2023Raising} and \cite{Liu2024Mixture} explored the use of CLIP-ViT models, showcasing strong generalization due to their pre-trained visual-world knowledge.

Additionally, multi-scale networks analyze images at multiple resolutions to capture both global and local features. \cite{Deng2023Diffusion} demonstrated the effectiveness of wavelet-based multi-scale networks for robust face forgery detection. Meanwhile, dual-stream networks with cross-attention have been proposed by \cite{Xi2023AI}, where separate branches analyze texture and low-frequency artifacts, showing superior performance over traditional methods.

CLIP-based detectors, which learn joint image-text representations, have also emerged as strong contenders. For instance, \cite{Santosh2024Robust} combined CLIP features with an MLP classifier, while \cite{Song2024Trinity} fused CLIP-extracted text features with pixel-level artifacts. These models demonstrate robust generalization across various detection tasks.

Other advanced approaches include autoencoder reconstruction error-based detection, which exploits the autoencoder component in diffusion models. For example, AEROBLADE \cite{Ricker2024AEROBLADE} is training-free, while DIRE \cite{Wang2023DIRE} uses a pre-trained diffusion model for reconstruction error analysis. \cite{Luo2024LaRE} further refined this approach with Latent Reconstruction Error (LaRE) for improved accuracy and efficiency.

Finally, methods analyzing intrinsic dimensionality and stepwise error analysis are gaining attention. \cite{Lorenz2023Detecting} employed multiLID for effective detection and generator identification, and \cite{Tulchinskii2023Intrinsic} explored its potential for text detection. Additionally, \cite{Ma2023Exposing} introduced SeDID, exploiting deterministic errors in diffusion models’ reverse and denoising processes, combining statistical and neural network approaches.

\subsection{Hybrid Approaches}

Hybrid methods that combine different analysis techniques are becoming increasingly popular. For instance, \cite{Song2024Trinity} effectively fused frequency and spatial information by combining text features, spectral analysis, and pixel-level artifact detection.

Integrating deep learning with statistical methods has also shown promise. \cite{Ma2023Exposing} combined statistical analysis with neural networks in SeDID, while \cite{Bammey2023Synthbuster} integrated Fourier analysis with a deep learning classifier, marking a trend towards combining data-driven and knowledge-driven approaches for more effective detection.

\section{Detection Methods Based on Textual and Multimodal Analysis for Text-to-Image Models}

With the increasing sophistication of text-to-image diffusion models, detecting AI-generated content requires a deep understanding of the relationships between input text prompts and generated images. Research in this area is growing rapidly, exploring approaches that utilize both textual and visual features to improve detection capabilities.

One approach focuses on analyzing the correlation between text prompts and their corresponding images. Several studies have examined how certain prompt characteristics can influence the realism of generated images. For example, \cite{Sha2022DE} systematically studied the effects of prompt topics and lengths on image authenticity, finding that certain prompt types, such as those centered around ``person," or prompts of specific lengths (e.g., 25-75 characters), led to more realistic images. These findings suggest that analyzing text prompts, including their topics, lengths, and even semantic nuances, can be an effective tool for distinguishing between AI-generated and authentic images. Similarly, \cite{Papa2023On} demonstrated the ability of prompts to generate highly realistic faces using Stable Diffusion v1.5, further underscoring the need to study the interplay between text and generated content.

Building on the correlation between text and image features, multimodal detection techniques are gaining popularity. These methods combine both textual and visual data, leveraging the complementary information found in each modality. \cite{Song2024Trinity} introduced the Trinity Detector, which integrates text features from a CLIP encoder with pixel-level artifacts. Their model, using a Multi-spectral Channel Attention Fusion Unit (MCAF), significantly improves detection performance by identifying subtle inconsistencies between the input prompt and the generated image. Additionally, \cite{Xu2023Exposing} presented a hybrid neural network that fuses attention-guided feature extraction with a vision transformer-based architecture, capturing both long-range and global image features. This multimodal approach demonstrates superior detection capabilities, emphasizing the importance of combining linguistic and visual analyses for both universal detection and source attribution.

Another promising direction in AI-generated image detection lies in identifying inconsistencies between the text prompt and the generated image. Authentic images typically exhibit strong semantic and structural alignment with their captions, while AI-generated images might show subtle discrepancies. While research in this area is still in its early stages, potential methods could include comparing semantic similarity between the prompt and image content using models like CLIP, or analyzing spatial relationships between objects described in the prompt and those depicted in the image. These inconsistencies can be particularly useful in cases where the generated image is highly realistic, and traditional artifact-based detection methods are less effective. Future research could explore how such mismatches evolve throughout the diffusion process, offering deeper insights into the generative mechanisms and potentially leading to more robust detection strategies.

Detecting AI-generated images from text-to-image models can benefit from a combination of textual analysis, multimodal detection methods, and the exploration of text-image inconsistencies. By leveraging insights from prompt characteristics, fusing textual and visual features, and examining the coherence between text and image, researchers can develop more effective detection methods for distinguishing AI-generated content from authentic images.

\section{Datasets and Benchmarks}

Evaluating the effectiveness of diffusion model-generated content detection requires robust and diverse datasets. Benchmark datasets serve as crucial tools in assessing the performance and generalizability of detection methods, ensuring detectors can handle various scenarios and challenges. This section reviews existing datasets used for this purpose and discusses the need for more diverse, challenging benchmarks to keep pace with rapidly advancing generative technologies.

\subsection{Existing Datasets for Evaluating Diffusion Model Detection}

Several datasets have been developed to test the robustness of AI-generated image detectors. These datasets vary in their scale, diversity, and the types of challenges they present, offering a broad spectrum for evaluating detection models.

One such dataset is \textbf{GenImage} \cite{Zhu2023GenImage}, a million-scale benchmark designed specifically to evaluate AI-generated image detectors. GenImage features over one million image pairs that cover a broad range of classes, including realistic degradations such as blurring and compression. This dataset is instrumental in testing detector performance across different generative models, including diffusion models and GANs. Its two primary evaluation tasks---cross-generator image classification and degraded image classification---provide valuable insights into how detectors perform when trained on one generator and tested on others, as well as how they handle low-quality images. This is particularly relevant given the findings of \cite{Corvi2023On}, which emphasized the importance of testing detectors under real-world social media conditions involving compression and resizing.

Another dataset, \textbf{COCOFake} \cite{Amoroso2023Parents}, offers a large-scale collection of around 1.2 million images generated from COCO image-caption pairs using Stable Diffusion v1.4 and v2.0. COCOFake is particularly useful for studying multimodal deepfake detection, as it links generated images with the captions used to create them. This allows researchers to explore how text prompts influence the characteristics and authenticity of generated images, aligning with the work in \cite{Sha2022DE}, which examined the interplay between text captions and image authenticity.

For facial forgery detection, the \textbf{DiFF} dataset \cite{Cheng2024Diffusion} provides a collection of over 500,000 fake facial images synthesized by thirteen different generation methods. These images are created under diverse conditions using 30,000 carefully curated textual and visual prompts, ensuring high fidelity and semantic consistency. The dataset is particularly well-suited for evaluating detectors in scenarios that mimic realistic facial forgery, which is becoming increasingly difficult to detect as AI-generated faces grow more realistic. As emphasized by \cite{Papa2023On}, the realism of AI-generated faces calls for detectors that remain robust under various image perturbations.

To test the generalizability of detectors, \textbf{WildFake} \cite{Hong2024WildFake} compiles a diverse range of fake images generated by various state-of-the-art models, including diffusion models, GANs, and other generative techniques. WildFake’s hierarchical structure, which organizes images by generator type, allows for a more targeted evaluation of detector performance. This dataset is particularly valuable for assessing how detectors generalize to unseen models and perform in real-world scenarios, where images can vary widely in class, style, and source, similar to the benchmark created in \cite{Wang2023DIRE}.

\subsection{The Need for More Diverse and Challenging Datasets}

While existing datasets like GenImage, COCOFake, DiFF, and WildFake provide a strong foundation for evaluating diffusion model detection methods, the rapid evolution of these models presents new challenges that current benchmarks may not adequately capture. There is a growing need for datasets that reflect a wider range of diffusion models, image transformations, and real-world conditions.

Current benchmarks tend to focus on a limited set of diffusion models. To fully evaluate the generalizability of detection methods, it is essential to develop datasets that encompass a broader spectrum of models, including both established and emerging architectures. This would help identify vulnerabilities specific to certain models and ensure detectors perform effectively across a variety of generative techniques, as suggested by \cite{Das2023Universal, peng2024jailbreaking}.

Moreover, real-world images often undergo various transformations and post-processing techniques, such as compression, resizing, filtering, and color adjustments. Datasets that include these types of image manipulations are critical for testing the robustness of detection methods under practical conditions. As discussed in \cite{Saberi2023Robustness}, images encountered on social media platforms are frequently degraded by compression or resizing, making it crucial for detectors to maintain accuracy despite these alterations. This need for robustness aligns with the challenges outlined in \cite{Lu2024Towards}, which emphasizes the importance of detectors that can handle image perturbations.

Finally, there is a growing need for datasets that reflect mixed real and synthetic content. In many real-world scenarios, images may contain both genuine and AI-generated elements, such as in the case of inpainting or manipulation. Datasets that feature this mixed-media reality are essential for evaluating the performance of detectors at a pixel level, ensuring they can distinguish between real and generated components within an image. As noted by \cite{Epstein2023Online}, detection methods need to be capable of operating in these complex, hybrid environments, pushing the boundaries of current detection capabilities. This challenge has been addressed in part by weakly supervised approaches like those described in \cite{Tantaru2024Weakly}, but more sophisticated datasets are needed to further drive advancements in this area.

\section{Evaluation Metrics}

When evaluating diffusion-generated content detectors, several metrics from traditional classification tasks and generative model assessments come into play. This section explores both the standard metrics used in classification tasks and those specific to generative models, while also considering the need for new metrics to address the unique challenges posed by diffusion models.

\subsection{Standard Classification Metrics}

The effectiveness of detectors for diffusion-generated content is often measured using standard classification metrics such as accuracy, precision, recall, F1-score, and AUROC (Area Under the Receiver Operating Characteristic curve). Accuracy provides an overall measure of the detector's correctness, while precision and recall respectively quantify the system's ability to minimize false positives (classifying real content as generated) and false negatives (failing to detect generated content). The F1-score, a harmonic mean of precision and recall, is widely used to balance these two aspects. AUROC assesses the detector's performance across various thresholds.

These metrics are commonly used in studies such as \cite{Bammey2023Synthbuster}, \cite{Ma2023Exposing}, and \cite{Das2023Universal}, with reported accuracies often exceeding 90\%. However, while these metrics are useful for general performance assessment, they provide limited insight into the nuanced challenges of detecting diffusion-generated content, especially regarding the quality, subtlety, and real-world impact of generated outputs. For instance, a detector may achieve high accuracy by exploiting easily detectable artifacts while struggling with more subtle manipulations \cite{Papa2023On}.

\subsection{Generative Model-Specific Metrics}

In addition to standard classification metrics, generative model-specific metrics like Fréchet Inception Distance (FID) and Inception Score (IS) offer a complementary perspective by quantifying the quality of generated images. FID measures the difference between the feature distributions of real and generated images, with a lower score indicating greater similarity. IS evaluates the quality and diversity of generated images. Both metrics have been widely adopted in evaluating generative models, though their relationship to detection performance remains complex.

For example, a low FID score suggests high-quality generative outputs, but these images may still contain detectable artifacts. \cite{Somepalli2023Diffusion} highlights how diffusion models sometimes replicate training data, which may artificially lower FID but potentially make detection easier. Moreover, emerging metrics like the Image Realism Score (IRS) \cite{Chen2023On} attempt to quantify the realism of images and distinguish between real and fake content, adding another dimension to the evaluation of diffusion models.

\subsection{Emerging Needs for New Metrics}

As diffusion models continue to evolve in complexity, new evaluation metrics are necessary to capture the specific attributes of their generated content. Existing metrics often fail to account for semantic consistency, such as the alignment between generated images and accompanying text prompts, which is crucial for text-to-image models \cite{Song2024Trinity}. Robustness to adversarial attacks and post-processing operations is another critical concern, particularly for real-world applications. \cite{Saberi2023Robustness} explores the vulnerability of detectors to various attacks, stressing the need for metrics that evaluate robustness and adversarial resistance.

Additionally, detection systems must consider application-specific contexts. For instance, the impact of generated content on human perception is crucial for assessing its real-world implications, as explored in \cite{Daphne2019Human}. Such factors underscore the need for more sophisticated and holistic evaluation frameworks that go beyond traditional metrics.

\section{Applications and Implications}

The detection of diffusion-generated content has far-reaching applications, from copyright protection to ethical considerations. Below, we explore some of the key areas where detection systems play a crucial role, along with their societal and legal implications.

\subsection{Copyright Protection and Content Authentication}

With diffusion models becoming increasingly sophisticated, protecting intellectual property rights is paramount. Diffusion-generated content can blur the lines between original artwork and AI-generated imitations, as seen in cases where models directly copy training data \cite{Somepalli2023Diffusion}. Techniques like watermarking, explored by \cite{Wen2023Tree} and \cite{Min2024Watermark}, aim to embed ownership information in generated content, allowing for subsequent detection and verification. However, ensuring the robustness of these techniques remains a challenge, especially in the face of watermark removal attacks \cite{Hu2024Stable}.

\subsection{Combating Misinformation and Deepfakes}

The rise of diffusion-generated deepfakes poses significant threats to online information integrity. Such synthetic content can be weaponized to spread misinformation, manipulate public opinion, or harm individual reputations. Detection methods are crucial for mitigating these risks by identifying and flagging manipulated or synthetic content. Research on human perception of deepfakes, such as \cite{Frank2023Representative}, also highlights the importance of understanding how realistic generated content can influence human judgment.

\subsection{Forensic Analysis and Investigation}

In forensic contexts, identifying the origin and authenticity of digital media is vital. Diffusion-generated content detection techniques provide tools for tracing manipulated or synthetic images back to their source. Methods like those proposed by \cite{Sinitsa2024Deep} focus on establishing relationships between fine-tuned generative models and the content they produce, which can aid in identifying the specific model used in a deepfake. Watermarking and fingerprinting techniques, discussed in \cite{Yu2021Artificial}, further enhance the ability to attribute generated content to its origin.

\subsection{Ethical Considerations and Responsible AI Development}

The ethical implications of diffusion models are broad and complex. As these models advance, their potential for misuse grows, whether in generating harmful content, violating copyright, or disseminating misinformation. Responsible AI development practices are essential to address these concerns. For instance, \cite{Gandikota2023Erasing} discusses methods for removing specific visual concepts from diffusion models to prevent undesirable outputs. In addition to detection methods, there is a growing consensus on the need for clear ethical guidelines and regulations. \cite{Knott2023Generative} argues for the mandatory implementation of detection mechanisms in publicly released generative models to ensure accountability and minimize harm.

\section{Research Gaps and Future Directions}

The ongoing development of diffusion models presents a range of challenges for detection methods. This section outlines key areas that require further research, from enhancing detection robustness to addressing ethical concerns.

\subsection{Enhancing Robustness and Generalization of Detection Methods}

Developing robust and generalizable detection methods for diffusion-generated content is a major challenge. Current detectors often fail to generalize across different diffusion models, datasets, and post-processing techniques. For example, \cite{Wang2020CNN} demonstrated that a classifier trained on a GAN model might generalize across GAN architectures but not to diffusion models. Similarly, \cite{Ojha2023Towards} highlighted the limitations of traditional deep network classifiers when applied to newer generative models. New approaches, such as leveraging frequency domain analysis \cite{T2019Fourier}, \cite{Lorenz2023Detecting}, adaptive learning algorithms, domain adaptation techniques \cite{Bhattacharjee2023ConDA}, and universal image and text representations, are promising but need further exploration.

\subsection{Using Multimodal and Cross-Modal Information for Detection}

With the increasing use of text-to-image diffusion models, integrating multimodal and cross-modal detection techniques becomes crucial. Most current detection approaches focus only on image analysis, but incorporating textual information could enhance detection accuracy. For instance, \cite{Xu2023Exposing} proposed a hybrid neural network combining attention and vision transformer components, while \cite{Song2024Trinity} fused text and pixel-level features. Future work should explore how to effectively integrate both text and image data using methods like cross-attention mechanisms or novel architectures. Additionally, analyzing prompts \cite{Sha2022DE} could offer insights into how text influences the detectability of generated images.

\subsection{Investigating the Impact of Training Data and Model Architectures}

The performance of detection methods is strongly influenced by the training data and model architecture. \cite{Somepalli2023Diffusion} showed the impact of dataset size and composition on replication rates, while \cite{Sinitsa2024Deep} demonstrated that certain CNN architectures could perform well even with limited training samples. Future research should examine how various data augmentation techniques \cite{Wang2020CNN}, dataset diversity, and detector architectures influence performance and generalization.

\subsection{Standardized Evaluation Metrics and Benchmarking}

Creating standardized evaluation metrics and benchmark datasets is essential for advancing detection methods. While existing datasets like \cite{Zhu2023GenImage} provide valuable resources, the rapidly evolving diffusion model landscape demands continuous updates. Future research should focus on expanding benchmark datasets to cover a diverse range of models, image resolutions, post-processing techniques, and real-world scenarios involving both synthetic and mixed real-synthetic content \cite{Epstein2023Online}. In addition, standardized evaluation protocols are needed to enable consistent and reproducible comparisons across detection methods.

\subsection{Ethical and Societal Implications of Diffusion-Generated Content}

The ethical concerns surrounding diffusion-generated content require careful attention. These models can be misused for creating deepfakes, spreading misinformation, and violating copyright, as highlighted by \cite{Corvi2023On}. Mandatory detection mechanisms, as advocated by \cite{Knott2023Generative}, are crucial to ensure responsible AI development. Future work should focus on developing ethical guidelines, promoting transparency in model releases, and raising public awareness about the risks and limitations of diffusion models.

\subsection{Adversarial Training and Defense Mechanisms}

The dynamic between generative models and detectors calls for advanced adversarial training and defense techniques. Research by \cite{Hooda2024D4} has shown that disjoint ensembles can improve robustness against adversarial attacks, while \cite{Saberi2023Robustness} analyzed detector vulnerabilities to sophisticated attacks like diffusion purification. Future efforts should explore novel adversarial training methods, build defenses against evolving attacks, and investigate the theoretical limits of robustness in diffusion model detection.

\subsection{Advances in Watermarking, Copyright Detection, and Backdoor Attack Prevention}

Watermarking, fingerprinting, and methods to detect disguised copyright infringement face growing challenges. Techniques like those proposed in \cite{Wen2023Tree} and \cite{Yu2021Artificial} for content authentication show promise, but attacks such as those discussed by \cite{Jiang2023Evading} highlight vulnerabilities. Similarly, detecting backdoor attacks on diffusion models is an ongoing concern, with research like \cite{Chou2023How} offering frameworks for backdoor detection and mitigation. Further studies should enhance watermark robustness, develop backdoor defense mechanisms, and explore advanced strategies for detecting copyright infringement \cite{Lu2024Disguised}.

\subsection{Role of Human Perception and Explainability}

Human perception plays a critical role in assessing diffusion-generated content. Studies such as \cite{Frank2023Representative} and \cite{Zeyu2023Seeing} suggest that people struggle to distinguish between real and AI-generated media, which raises concerns about the potential for misinformation. Research should investigate cognitive biases, cross-cultural differences in perception, and strategies for improving human detection abilities. At the same time, the explainability of detection models is essential for building trust and transparency. Techniques such as Layer-wise Relevance Propagation, as explored in \cite{Guo2023Accurate}, and attention mechanisms should be further developed to provide human-understandable justifications for detection decisions.

\subsection{Exploring Positive Applications of Diffusion Models}

In addition to detection, diffusion models have potential benefits in various fields. For instance, \cite{Chen2024Synthetic} used diffusion models to augment weed identification data, while \cite{Wu2023DatasetDM} generated synthetic datasets with perception annotations. Future research should focus on exploring the use of diffusion models to generate synthetic data in fields like medical imaging \cite{Ali2022Spot}, material science, and robotics, where high-quality data is often scarce.

\subsection{Advancements in Specialized Domains}

Diffusion models offer potential advancements in several specialized domains. For example, generating synthetic medical images with higher fidelity is a key area of research \cite{Ali2022Spot}. Conditional generation techniques, anatomical constraints, and robust evaluation metrics should be explored to improve the quality of these images. Similarly, diffusion models can be used for camouflaged object detection (COD), as demonstrated by \cite{Luo2023CamDiff}, to synthesize challenging datasets for training COD models. Exploring adversarial examples for COD models could also help enhance their robustness.

\bibliographystyle{IEEEtran}  
\bibliography{references}

\end{document}